\documentclass[11pt,letterpaper]{article}
\usepackage{emnlp2016}
\usepackage{times}
\usepackage{multirow}
\usepackage{latexsym}
\usepackage{mathrsfs}
\usepackage{graphicx}
\usepackage{subfig}
\usepackage{amsfonts}
\usepackage{amssymb}
\usepackage{amsmath}
\usepackage{bm}
\usepackage{url}
\usepackage{tikz-dependency}
\usepackage{pgfplots}
\usepackage{xcolor}
\usepackage{array}

\usepackage{booktabs}

\emnlpfinalcopy 
\allowdisplaybreaks

\newenvironment{itemize*}%
  {\begin{itemize}%
    \setlength{\itemsep}{0pt}%
    \setlength{\parskip}{0pt}}%
  {\end{itemize}}
  \newenvironment{enumerate*}%
  {\begin{enumerate}%
    \setlength{\itemsep}{0pt}%
    \setlength{\parskip}{0pt}}%
  {\end{enumerate}}

\newcommand{\snli}[3]{{\vspace{0.25em}
{ \setlength{\parindent}{0.6em} \hangindent=1.2em  \textbf{Premise:} #1\par}\vspace{0.25em}
{ \setlength{\parindent}{0.6em} \hangindent=1.2em   \textbf{Hypothesis:} #2\par}\vspace{0.25em}
{ \setlength{\parindent}{0.6em}  \textbf{Label:} #3\par}
}}

\DeclareMathOperator*{\softmax}{softmax}
\def\W{\mathbf{W}}
\def\U{\mathbf{U}}
\def\bb{\mathbf{b}}

\def\h{\mathbf{h}}
\def\rr{\mathbf{r}}
\def\bx{\mathbf{x}}

\def\cc{\mathbf{c}}
\def\ii{\mathbf{i}}
\def\ff{\mathbf{f}}
\def\oo{\mathbf{o}}

\def\cc{\mathbf{c}}

\def\R{\mathbb{R}}

\def\w{\mathbf{w}}

\title{Syntax-based Attention Model for Natural Language Inference}
\author{Pengfei Liu \quad Xipeng Qiu\thanks{Corresponding author.} \quad Xuanjing Huang\\
Shanghai Key Laboratory of Intelligent Information Processing, Fudan University\\
School of Computer Science, Fudan University\\
825 Zhangheng Road, Shanghai, China\\
\{pfliu14,xpqiu,xjhuang\}@fudan.edu.cn}

\begin{document}
\maketitle
\begin{abstract}
Introducing attentional mechanism in neural network is a powerful concept, and has achieved impressive results in many natural language processing tasks. However, most of the existing models impose attentional distribution on a flat topology, namely the entire input representation sequence. Clearly, any well-formed sentence has its accompanying syntactic tree structure, which is a much rich topology. Applying attention to such topology not only exploits the underlying syntax, but also makes attention more interpretable. In this paper, we explore this direction in the context of natural language inference. The results demonstrate its efficacy. We also perform extensive qualitative analysis, deriving insights and intuitions of why and how our model works.

\end{abstract}


\section{Introduction}

Recently, adopting neural attentional mechanism has proven to be an extremely successful technique in a wide range of natural language processing tasks, ranging from  machine translation \cite{bahdanau2014neural}, sentence summarization \cite{rush2015neural}, question answering \cite{hermann2015teaching} and text entailment \cite{rocktaschel2015reasoning,wang2015learning,cheng2016long}. 
The basic idea is to learn and attend to most relevant parts of (potentially preprocessed) a sequence $X$ while analysing or generating another sequence $Y$.

\begin{figure}[t]\centering
  \includegraphics[width=0.9\linewidth]{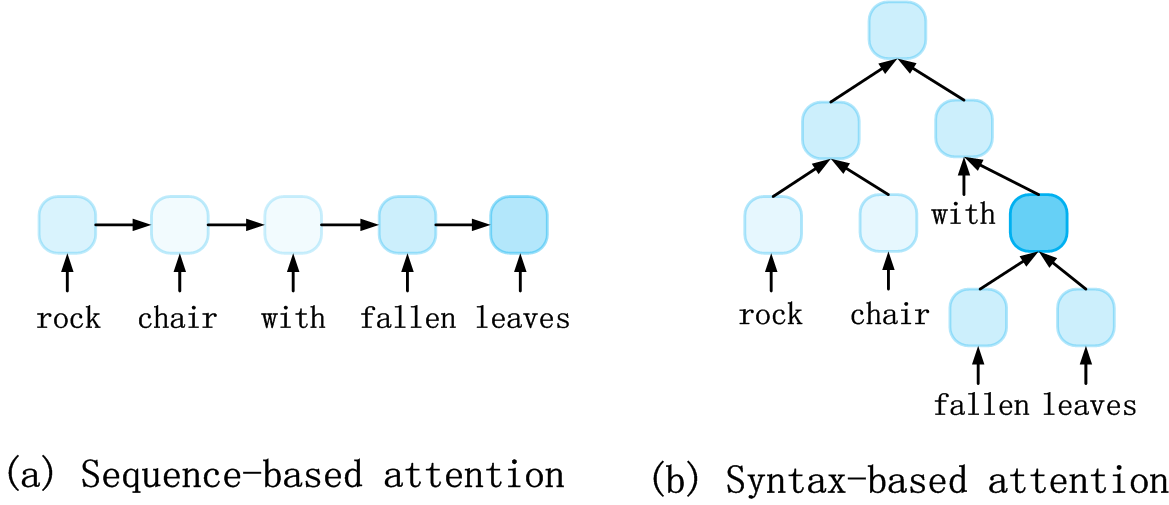}
  \caption{A motivated example to illustrate sequence-based and syntax-based attention model for target word ``\texttt{autumn}''.
  The square boxes represent hidden states of the words (or phrases); darker indicates higher alignment.
  }\label{fig:example}
\end{figure}

Taking the following two sentences as examples, where we highlight the helpful partial information alignment from $X$ according to $Y$ with attention.\\
$X$: \textit{A toddler sits on a rock chair with {\bf fallen leaves}.}\\
$Y$: \textit{A little child sits quietly on a stone bench in {\bf autumn}.}

The sequence-based attention is illustrated in Figure \ref{fig:example}(a). The representation is a flat sequence, and attention distribution is applied to this simple topology. Although the idea is to soft-align words and phrases in the two sentences, one can observe that:
1) The hidden state of each position incorporates its context information, which is implicit and sequential, alignment at phrase-level is thus challenging (e.g. ``autumn'' to ``fallen leaves'').  2) As we will discuss shortly, the attention is implemented with a weighted sum of sequence, thus lacks linguistic interpretation for its semantic composition.




Any well-formed sentences have its underlying syntactic structure. It is a tree topology that encodes a sentence's important composing subcomponents. Evidently, this is in stark contrast with the flat and sequential topology the existing models assume.

In this paper we extend the attentional mechanism from a sequence to a tree, allowing syntactic information to be integrated.
As shown in Figure \ref{fig:example}(b), syntax-based attention allows neural models to more explicitly capture the phrase-level alignment. In addition, it clearly reaches a higher level of interpretability. 
While this observation is general, in this paper we demonstrate its effectiveness in natural language inference. We believe other tasks such as neural translation model \cite{bahdanau2014neural,luong-pham-manning:2015:EMNLP} can similarly benefit from this idea.

 The contributions of this paper can be summarized as follows.
 \begin{enumerate}
    \item We extend sequence-based attention to syntax-based, therefore incorporating richer linguistic properties.
    \item We design and validate our algorithm that makes such topological attentional mechanism possible.
    \item Beyond quantitative measurement, we carefully perform qualitative analysis, and demonstrate why and how the idea works.
    \item Our work can be regarded as an attempt to boost the generalization ability of attention matching mechanism by encoding prior knowledge (syntax). As an example, our results show syntactic structure of sentence or phrase is crucial for text semantic matching.
 \end{enumerate}

\section{Neural Attention Model for Natural Language Inference}
Natural language inference, also called text entailment, is a task to determine the semantic relationship (entailment, contradiction, or neutral) between two sentences (a premise and a hypothesis). 
This task is important involved in many natural language processing (NLP) problems, such as information extraction, relation extraction, text summarization or machine translation.

To better understand this task, we give an example in the dataset as follows:

\snli{These girls are having a great time looking for seashells.}
{The girls are happy.}
{entailment}

More precisely, NLI can be framed as a simple three-way classification task, which requires the model to be able to represent and reason with the core phenomena of natural language semantics \cite{bowman2016fast}.

\subsection{Long Short-Term Memory Network}

Long short-term memory neural network (LSTM)~\cite{hochreiter1997long} is a type of recurrent neural network (RNN) \cite{Elman:1990}, and
specifically addresses the issue of learning long-term dependencies. LSTM maintains a memory cell that updates and exposes its content only when deemed necessary.

While there are numerous LSTM variants, here we use the LSTM architecture used by \cite{jozefowicz2015empirical}, which is similar to the one in \cite{graves2013generating} but without peep-hole connections.

We define the LSTM \emph{units} at each time step $t$ to be a collection of vectors in $\mathbb{R}^d$: an \emph{input gate} $\ii_t$, a \emph{forget gate} $\ff_t$,  an \emph{output gate} $\oo_t$, a \emph{memory cell} $\cc_t$ and a hidden state $\h_t$. $d$ is the number of the LSTM units. The elements of the gating vectors $\ii_t$, $\ff_t$ and $\oo_t$ are in $[0, 1]$.

The LSTM is precisely specified as follows.

\begin{align}
	\begin{bmatrix}
		\mathbf{\tilde{c}}_{t} \\
		\mathbf{o}_{t} \\
		\mathbf{i}_{t} \\
		\mathbf{f}_{t}
	\end{bmatrix}
	&=
	\begin{bmatrix}
		\tanh \\
		\sigma \\
		\sigma \\
		\sigma
	\end{bmatrix}
	T_{\mathbf{A},\mathbf{b}}
	\begin{bmatrix}
		\mathbf{x}_{t} \\
		\mathbf{h}_{t-1}
	\end{bmatrix},\label{eq:lstm1}\\
\mathbf{c}_{t} &=
		\mathbf{\tilde{c}}_{t} \odot \mathbf{i}_{t}
		+ \mathbf{c}_{t-1} \odot \mathbf{f}_{t}, \\
	\mathbf{h}_{t} &= \mathbf{o}_{t}  \odot \tanh\left( \mathbf{c}_{t}  \right)\label{eq:lstm2},
\end{align}
where $\bx_t$ is the input at the current time step;
$T_{\mathbf{A},\mathbf{b}}$ is an affine transformation which depends on parameters of the network $\mathbf{A}$ and $\mathbf{b}$.
$\sigma$ denotes the logistic sigmoid function and $\odot$ denotes elementwise multiplication. 

The update of each LSTM unit can be written precisely as
\begin{align}
(\h_t,\cc_t) &= \mathbf{LSTM}(\h_{t-1},\cc_{t-1},\mathbf{x}_t).
\end{align}

Here, the function $\mathbf{LSTM}(\cdot, \cdot, \cdot)$ is a shorthand for Eq. (\ref{eq:lstm1}-\ref{eq:lstm2}).

LSTM can map the input sequence of arbitrary length to a fixed-sized vector, and has been successfully applied to a wide range of NLP tasks, such as machine translation \cite{sutskever2014sequence}, language modelling \cite{sutskever2011generating} and  natural language inference \cite{rocktaschel2015reasoning}.

\subsection{Neural Attention Model}

Given two sequences $X = x_1,x_2,\cdots,x_n$ and $Y = y_1,y_2,\cdots,y_m$, we let $\bx_i \in \R^d$ denote the embedded representation of the word $x_i$. The standard LSTM has one temporal dimension:
at position $i$ of sentence $x_{1:n}$, the output $\h^{x}_i$ reflects the meaning of the subsequence $x_{1:i}={x_1,\cdots,x_i}$.

The main idea of attention model \cite{hermann2015teaching} is that the representation of sentence $X$ is obtained dynamically based on the degree of alignment between the words in sentence $X$ and $Y$.
More formally, for sentence $X$ and $Y$, we first compute the hidden state of each sentence by two LSTMs: 
\footnote{The model used by \cite{rocktaschel2015reasoning} is a little different from this for a better performance, in which encoding of one sentence is conditioned on the other.}:
\begin{align}
\h_i^{x} &= \mathbf{LSTM}(\h_{i-1}^{x},\cc_{i-1}^{x},\mathbf{x}_i) \\
\h_j^{y} &= \mathbf{LSTM}(\h_{j-1}^{y},\cc_{j-1}^{y},\mathbf{y}_j)
\end{align}
While processing sentence $Y$ at time $j$, the model emits an attention vector $\alpha_j \in \R^n$ to weight $\h_i^x$ , the hidden states of $X$, thereby obtaining a fine-grained representation $\mathbf{r}$ of sentence $X$ as follows:
\begin{align}
    \mathbf{r}_j^{x} = \sum_{i=1}^n{\alpha_{ji}\h_i^x} + \tanh(\W^{r}\mathbf{r}^x_{j-1}) \label{eq:att1}
\end{align}
where $\alpha_{ji}$ can be compute as:
\begin{align}
\alpha_{ji} = softmax(e_{ji})=\frac{\exp(e_{ji})}{ \sum_{i'}{\exp(e_{ji'})}} \label{eq:att2}
\end{align}
Where $e_{ji}$ is a alignment score and can obtained by:
\begin{align}
e_{ji} = \w^e \cdot \tanh(\W^y\h_{j}^{y} + \W^x\h_{i}^{x} + \W^r \mathbf{r}_{j-1}^x) \label{eq:att3}
\end{align}
where $\W^y$, $\W^x$, $\W^r$ are learned parameters.

Finally, the representation of the sentence pair $\h^{\ast}$ is constructed by the last attention-weighted representation $\mathbf{r}_m$ and the last output vector $\h_{m}^y$ as:
\begin{align}
\h^{\ast} = \tanh(\W^x \mathbf{r}_m^x + \W^y \h_{m}^{y} ). \label{eq:att4}
\end{align}

\begin{figure}[t]\centering

  \includegraphics[width=1\linewidth]{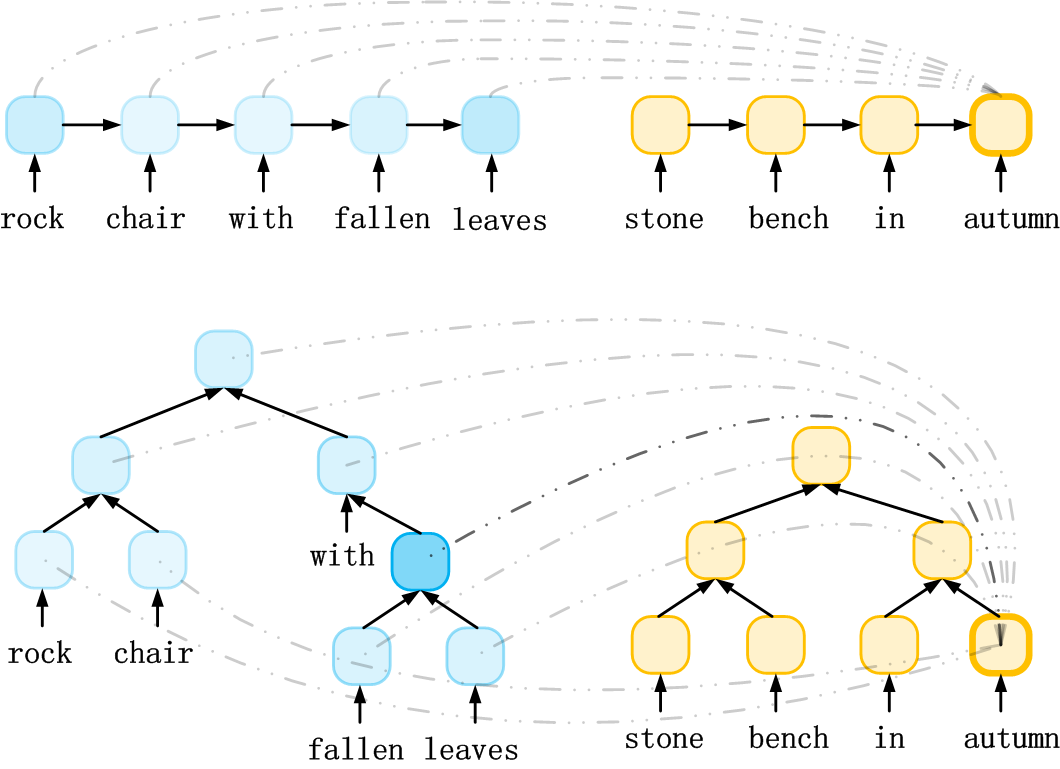}

  \caption{Two matching frameworks: Sequence-based attention model and syntax-based attention model.
  The box represents hidden state $\h$ of a node and the bold yellow box represents the node $y_j$ of sentence $Y$ at the position $j$.
  The darker blue box represents a higher alignment score between the corresponding node and the node $y_j$.
  }\label{fig:two-framework}
\end{figure}

For the entailment task, the final representation $\h^{\ast}$ of sentence-pair, is fed into the output layer, generating the probabilities over all pre-defined classes (entailment, contradiction, or neutral)  .
\begin{align}
{\hat{\mathbf{l}}} = \softmax(\W^{o} \h^{\ast} + \bb^{o})
\end{align}
where $\W^{o}$ and $\bb^{o}$ are parameters of the model.

\section{Syntax-Based Attention Matching Model}

The building block of this work syntax-based instead of sequence-based compositional model. There are several such candidates, such as recursive neural network \cite{socher2013recursive} and tree-structured LSTM \cite{tai2015improved}. In this paper, we use latter model since for its superior performance in representing sentence meaning.

\subsection{Tree-structured LSTM}


Different with standard LSTM, tree-structured LSTM composes its state from an input vector and the hidden states of children units. More formally, the model takes as input a syntactic tree (constituency tree or dependency tree), then a composition function is applied to combine the children nodes according to the syntactic structure to obtain an new compositional vector for their parent node. 

Here we investigate two types of composition functions for constituency and dependency tree respectively.

\paragraph{Composition Function for Constituency Tree}

Given constituency tree $T$ induced by a sentence, there are at most  $N$ children nodes for each parent node. We refer to $\h_{jk}$ and $\cc_{jk}$ as the hidden state and memory cell of the $k$-th child of node $j$.   The transition equations of each node $j$ are as follows:


\begin{align}
	\begin{bmatrix}
		\mathbf{\tilde{c}}_{j} \\
		\mathbf{o}_{j} \\
		\mathbf{i}_{j}
	\end{bmatrix}
	&=
	\begin{bmatrix}
		\tanh \\
		\sigma \\
		\sigma
	\end{bmatrix}
    \begin{pmatrix}
	\W^p
	\begin{bmatrix}
		\mathbf{x}_{j} \\
		\mathcal{H}_{j}
	\end{bmatrix}+\bb_p
    \end{pmatrix}, \label{eq:ctreelstm1}\\
\mathbf{f}_{jk} &= \sigma(\W^{f} \mathbf{x}_j + \U^{f}_k \mathcal{H}_j), \\
\mathbf{c}_{j} &=
		\mathbf{\tilde{c}}_{j} \odot \mathbf{i}_{j}
		+ \sum_{k}^{N} \mathbf{c}_{jk} \odot \mathbf{f}_{jk}, \\
	\mathbf{h}_{j} &= \mathbf{o}_{j}  \odot \tanh\left( \mathbf{c}_{j}  \right),\\
    \mathcal{H}_{j} &= \h_{j1}\oplus \h_{j2} \oplus  \cdots \oplus  \h_{jN},\label{eq:ctreelstm2}
\end{align}
where $\mathbf{x_j}$ denotes the input vector and is non-zero if and only if it is a leaf node.
$\sigma$ represents the logistic sigmoid function and $\odot$ denotes element-wise multiplication.
$\W^p$, $\W^f$, and $\U^k$ is the weight matrix which depends on parameters of the network.

\paragraph{Composition Function for Dependency Tree}
For the dependency tree, we refer to $C(j)$ as the set of children of node $j$.
Then the transition equations of each node $j$ are formulated as:

\begin{align}
	\begin{bmatrix}
		\mathbf{\tilde{c}}_{j} \\
		\mathbf{o}_{j} \\
		\mathbf{i}_{j}
	\end{bmatrix}
	&=
	\begin{bmatrix}
		\tanh \\
		\sigma \\
		\sigma
	\end{bmatrix}
    \begin{pmatrix}
	\W^p
	\begin{bmatrix}
		\mathbf{x}_{j} \\
		\tilde{\mathcal{H}}_{j}
	\end{bmatrix}+\bb_p
    \end{pmatrix}, \label{eq:dtreelstm1}\\
\mathbf{f}_{jk} &= \sigma(\W^{f} \mathbf{x}_j + \U^{f} \h_{jk}) \\
\mathbf{c}_{j} &=
		\mathbf{\tilde{c}}_{j} \odot \mathbf{i}_{j}
		+ \sum_{k}^{C(j)} \mathbf{c}_{jk} \odot \mathbf{f}_{jk}, \\
	\mathbf{h}_{j} &= \mathbf{o}_{j}  \odot \tanh\left( \mathbf{c}_{j}  \right)\label{eq:dtreelstm2},\\
    \tilde{\mathcal{H}}_{j} &= \sum_{k}^{C(j)} \h_{jk}
\end{align}
where  $\W^p$, $\W^f$, and $\U^f$ are the weight matrices which depend on parameters of the network.

The update of each unit can be written precisely as
\begin{align}
\h_j &= \mathbf{TreeLSTM}(\mathcal{H}_{j}, \tilde{\mathcal{H}}_{j}, \mathbf{x}_j).
\end{align}

Here, the function $\mathbf{TreeLSTM}(\cdot, \cdot, \cdot)$ is a shorthand for Eq. (\ref{eq:ctreelstm1}-\ref{eq:ctreelstm2}) for constituency tree or  (\ref{eq:dtreelstm1}-\ref{eq:dtreelstm2}) for dependency tree.

\subsection{Syntax-Based Attention Matching Model}
The second stage of the design is to apply attention to the tree topology. For two trees $T^{x}$ and $T^{y}$ induced by sentence $X$ and $Y$,  the representation of their subtrees $\h_i^{x}$ and $\h_j^{y} $ can be obtained as follows:
\begin{align}
    \h_i^{x} &= \mathbf{TreeLSTM}(\mathcal{H}_{i}^{x},\tilde{\mathcal{H}}_{j},\mathbf{x}_i) \\
    \h_j^{y} &= \mathbf{TreeLSTM}(\mathcal{H}_{j}^{y},\tilde{\mathcal{H}}_{j},\mathbf{y}_j)
\end{align}

At node $j$ of tree $T^{y}$, we reread over tree $T^{x}$ and compute a weighted tree representation $\mathbf{r}_j^{x}$  of  tree $T^{x}$, which also recursively accumulate information from its children $ \mathcal{R}_{j} = \{\rr_{j1},\rr_{j2}, \cdots, \rr_{jN}\}$.
\begin{align}
    \mathbf{r}_j^{x} &= \sum_{i=1}^{T_n}{\alpha_{ji}\h_i^x} +  \tanh( g(\mathcal{R}_{j}))
\end{align}
where $T_n$ denotes the number of nodes of tree $T_x$; $\alpha_{ji}$ measures the alignment degree between two subtrees; $g(\mathcal{R}_{j})$ is recursively accumulate information from its children.

For constituency tree,
\begin{align}
    g(\mathcal{R}_{j}) &= \W^{r} (\rr_{j1} \oplus \rr_{j2} \oplus  \cdots \oplus  \rr_{jN}).
\end{align}
For dependency tree,
\begin{align}
    g(\mathcal{R}_{j}) &= \tilde{\W}^{r} \sum_{k}^{C(j)} \rr_{jk}.
\end{align}

The attention $\alpha_{ji}$ between two subtrees $\h_{j}^{y}$ and $\h_{i}^{x}$ can be computed as
\begin{align}
e_{ji}    &= \w^e \cdot \tanh(\W^y\h_{j}^{y} + \W^x\h_{i}^{x} + g(\mathcal{R}_{j})),\\
  \alpha_{ji} &= softmax(e_{ji}).
\end{align}

The final representation $\h^{\ast}$ of two trees $T^{x}$ and $T^{y}$ can be obtained by
\begin{align}
\h^{\ast} &= \tanh(\W^x \mathbf{r}_{T_m}^x + \W^y \h_{T_m}^{y} ),
\end{align}
where $T_m$ denotes the number of nodes of tree $T_y$.

To facilitate the description later, we refer to SAT-LSTMs as our proposed syntax-based attention model. dLSTM and cLSTM represent LSTMs are built over a dependency and constituency respectively.

\section{Training}

 Given a sentence pair $(X,Y)$ and its label $l$. The output $\hat{l}$ of neural network is the probabilities of the different classes. The parameters of the network are trained to minimise the cross-entropy of the predicted and true label distributions.

\begin{equation}
  L(X,Y; \emph{\textbf{l}}, \hat{\emph{\textbf{l}}}) = - \sum_{j=1}^C  \emph{\textbf{l}}_j \log(\hat{\emph{\textbf{l}}}_j),
\end{equation}
where $\emph{\textbf{l}}$ is one-hot representation of the ground-truth label $l$; ${\hat{\emph{\textbf{l}}}}$ is predicted probabilities of labels; $C$ is the class number.

To minimize the objective, we use stochastic gradient descent with the diagonal variant of AdaGrad \cite{duchi2011adaptive}.
To prevent exploding gradients, we perform gradient clipping by scaling the gradient when the norm exceeds a threshold \cite{graves2013generating}.

\subsection{Initialization and Hyperparameters}

\paragraph{Orthogonal Initialization}
We use orthogonal initialization of our LSTMs, which allows neurons  to react to the diverse patterns and is helpful to train a multi-layer network \cite{saxe2013exact}.

\paragraph{Unsupervised Initialization}
The word embeddings for all of the models are initialized with the 100d GloVe vectors (840B token version, \cite{pennington2014glove}).
The other parameters are initialized by randomly sampling from uniform distribution in $[-0.1, 0.1]$.

\paragraph{Hyperparameters}
For each task, we take the hyperparameters which achieve the best performance on the development set via an small grid search over combinations of the initial learning rate $[0.05, 0.0005, 0.0001]$, $l_2$ regularization $[0.0, 5E{-5}, 1E{-5}, 1E{-6}]$ and the threshold value of gradient norm $\rho$ [5, 10, 50].
The final hyper-parameters are reported in Table \ref{tab:paramSet}.

\begin{table}[t]  \setlength{\tabcolsep}{3pt}
\centering
\begin{tabular}{l*{1}{p{0.18\linewidth}}}
    \toprule
    Hyper-parameters & SNLI\\ \hline
    Embedding size &  100\\
    Hidden layer size  & 100\\
    Initial learning rate & 0.005\\
    Regularization  & $0.0$\\
    $\rho$  & $50$\\

    \bottomrule
\end{tabular}
\caption{Hyper-parameters for our model on SNLI.}\label{tab:paramSet}
\end{table}

\section{Experiment}
We use the Stanford Natural Language Inference Corpus (SNLI) \cite{bowman-EtAl:2015:EMNLP}. This corpus contains 570K sentence pairs, and all of the sentences and labels stem from human annotators. SNLI is two orders of magnitude larger than all other existing RTE corpora. Therefore, the massive scale of SNLI allows us to train powerful neural networks such as our proposed architecture in this paper.

\subsection{Data Preparation}
We parse the sentences in the dataset for our tree-structured LSTMs. More specifically,
for the Dependency Tree-LSTMs, we produce dependency parses\cite{chen2014fast} of each sentence;  For constituency Tree-LSTMs, the trees are parsed by binarized constituency parser\cite{klein2003accurate}.

\subsection{Competitor Methods}
\begin{itemize}
  \item Neural bag-of-words (NBOW): Each sequence is represented as the sum of the embeddings of the words it contains, and then they are concatenated and fed to a multi-layer perceptron (MLP).
  \item LSTM encoders: The sentence pair are encoded by LSTMs respectively.

  \item Attention LSTM encoders (AT-LSTMs): The sentence pair are encoded with the consideration of the alignment of words between two sentences \cite{rocktaschel2015reasoning}.
  \item Tree-based CNN encoders:  The sentence pair are encoded by tree-based CNNs respectively \cite{mou2015recognizing}.
  \item Tree-based LSTM encoders:  The sentence pair are encoded by tree-based LSTM respectively.
  \item SPINN-PI encoder:  The sentence pair are encoded by stack-augmented parser-interpreter neural network with parsed input respectively, which is proposed by \cite{bowman2016fast}.
\end{itemize}

\begin{table*}[t]
  \centering \small
  \begin{tabular}{lrrrr}
\toprule
Model                   & Hidden.    & Train acc. (\%)  &   Dev. acc. (\%)  &   Test acc. (\%) \\
\midrule
\multicolumn{5}{c}{\textbf{Previous non-NN results}}\\
Lexicalized classifier \cite{bowman-EtAl:2015:EMNLP}
                        & ---                & 99.7                    & ---      &   78.2      \\
\midrule
\multicolumn{5}{c}{\textbf{Previous sentence encoder-based NN results}}\\
LSTM encoders \cite{bowman-EtAl:2015:EMNLP}
                        & 100                  & 84.8               &  ---    &   77.6      \\
Tree-based CNN encoders \cite{mou2015recognizing}
                        & 300                  & 83.4             & 82.4     &   82.1       \\
SPINN-PI encoders \cite{bowman2016fast}
                        & 300                  & 89.2                & ---        &   83.2       \\
AT-LSTMs encoders \cite{rocktaschel2015reasoning}
                        & 100                  & 85.3                &   83.7      &   \textbf{83.5}       \\

\midrule
\multicolumn{5}{c}{\textbf{Our results}}\\
Tree-dLSTM encoders
                        & 100                  & 83.5               &   77.1   &   78.7      \\
Tree-cLSTM encoders
                        & 100                  & 82.2               &   79.8   &   80.3      \\
AT-LSTMs encoders
                             & 100             & 84.2                &   82.7      &   82.0       \\
SAT-dLSTMs         & 100             & 86.6                &   83.8      &   83.4       \\
SAT-cLSTMs       & 100             & 87.9                &   85.0      &   \textbf{84.1}       \\

\bottomrule
  \end{tabular}
\caption{\protect\label{tab:results} Results of our proposed models against
other neural models on SNLI corpus.  Hidden. is the number of neurons in hidden state $\h$. Train, Dev. and Test denote the classification accuracy.
SAT-LSTMs denote our proposed syntax-based attention model. dLSTM and cLSTM represent LSTMs are built over a dependency and constituency respectively.  }
\end{table*}

\subsection{Results}
Table \ref{tab:results} provides a comparison of results on SNLI dataset.
From the table, we can observe that:
\begin{itemize}
    \item For two kinds of syntax-based LSTM encoders, cLSTM achieve better performances than  dLSTM, which is consistent with \newcite{gildea2004dependencies} experiment results on tree-based alignment. We think the reason is that constituency-based model can better learn the semantic compositionality and it has taken the orders of child nodes into consideration.
    \item Irrespective of attention mechanism, both two syntax-based LSTM encoders are superior to sequence-based LSTM encoder, which indicates the effectiveness of syntax-based composition.
    \item SAT-cLSTMs surpass all the competitor methods and achieve the best performance. More precisely, SAT-cLSTMs outperform AT-LSTMs by 2.1\%, and are superior to Tree-LSTM encoders by 3.8\%, which suggests the importance of incorporating syntactic information into attention models.
\end{itemize}

\begin{figure}[t]\centering

  \includegraphics[width=0.8\linewidth]{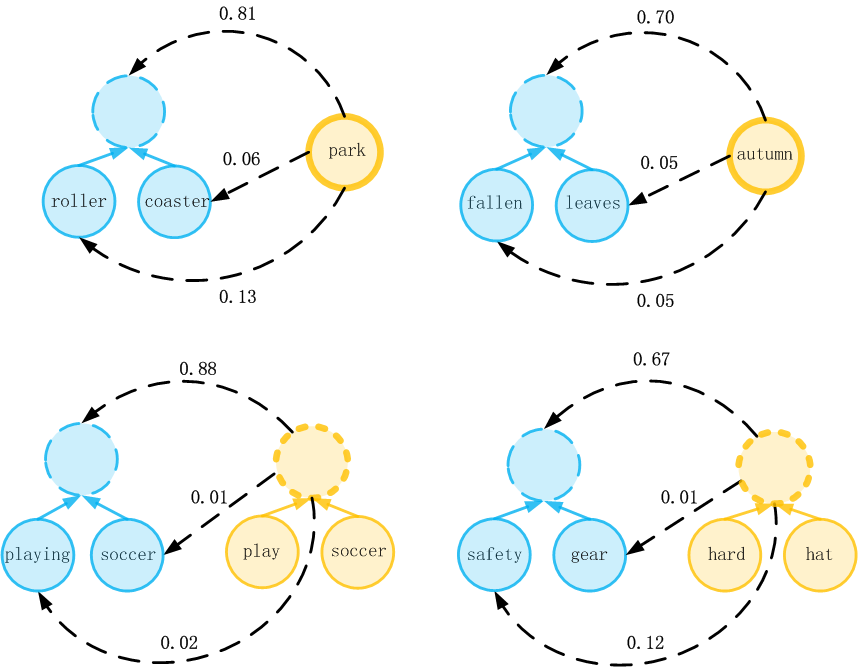}

  \caption{Visualization of syntax-based alignments over two subtrees. The numbers along the dotted lines represent the alignment scores.}\label{fig:example1}
\end{figure}

\subsection{Experiment Analysis}

\subsubsection{Analysis of Compositionality and Attention Mechanism}
Can our model select useful composition information using attention mechanism ?
To answer this question, we sample several sub-tree pairs from test dataset which achieve the best alignment of a sentence pair.

As shown in Figure \ref{fig:example1}, we can observe that,
\begin{itemize}
    \item The alignments in these cases are consistent with people's understanding. For example, the alignment degree $\alpha(autumn,fallen\   leaves)$ is much higher than $\alpha(autumn,fallen)$ and $\alpha(autumn,leaves)$, which is crucial for the final prediction of the two sentence' relation and indicates the effectiveness of this syntax-based composition.
    \item Our model has learned the alignment between subtrees, meaning that matching patterns at word-phrase or phrase-phrase level can be captured effectively not merely at word-word level.
\end{itemize}

\begin{table*}
\centering
\begin{tabular}{cccc}
\toprule
 \textbf{person 's} & \textbf{holding his cup up}  & \textbf{wearing a pink dress} & \textbf{having a great time} \\
\midrule
 people 's           & holding up a white plastic cup & in a pink dress      & having a good time\\
 belong to the lady  & with a cup in his hand         & dressed in pink      & enjoy time together \\
 of a person         & with a beer in his hand        & wearing a pink dress & is very happy\\
 of humans           & holds up a playing card        & in pink              & enjoying a night\\
\bottomrule
\end{tabular}
\caption{Nearest neighbor phrases drawn from the SNLI test set, which based on cosine similarity of different representations produced by SAT-LSTMs.}
\label{tab:nn_subtree}
\end{table*}

\begin{table*}[!t]
\center
\footnotesize
\em
\begin{tabular}{ccc}
\toprule
\addlinespace
& the boys are bare chested  & a golden retriever nurses puppies  \\
\cmidrule(lr){2-3}
\multirow{4}{*}{{\em {\normalsize NBOW}}}
& the men are naked         & a cat nurses puppies  \\
& the boys are stretching   &  a puppy barks at a girl   \\
& the boys are sleeping     & the dog is a labrador retriever \\
& the boys are sitting down & a golden retriever nurses some other dogs puppies \\

\addlinespace
\addlinespace
& the man has nothing on his face  & a girl is sitting on a park bench holding a puppy  \\
{\em {\normalsize AT-LSTMs}} 	
& a man is outside with no bag on his back  & a big dog watching over a smaller dog  \\
{\em {}}	& his bald head is exposed & the big dog is checking out the smaller dog  \\
			& a man in summer clothing skiing on thin snow  & a gal is holding a stuffed dog  \\
\addlinespace
\addlinespace
			& the man is not wearing a shirt& a golden retriever nurses some other dogs puppies \\
{\em {\normalsize SAT-LSTMs}}	
& two men are shirtless &  three puppies are snuggling with their mother by the fire   \\
{\em {}}	& the man is completely nude  & puppies next to their mother  \\
			& a man without a shirt is on the water & a mother dog checking up on her baby puppy   \\
\bottomrule
\end{tabular}
\caption{Nearest neighbor sentences drawn from the SNLI test set, which based on cosine similarity of different representations emitted by NBOW, AT-LSTMs and SAT-LSTMs.}
\label{tab:nn}
\end{table*}
\subsubsection{Analysis of Phrases Representations}
We compute the representations of each subtree and show some examples sampled from test dataset with their most related neighbors in Table \ref{tab:nn_subtree}.

The phrasal paraphrases, such as ``\texttt{having a great time/enjoy time together}'', have obtained close representations, which is more helpful for the identification of the entailment relation of two sentences. Besides, we can see the ability of the model to learn a variety of general paraphrastic transformations, such as possessive rule ``\texttt{persons's/of a person}'' and verb particle shift ``\texttt{holding his cup up/holding up a white plastic cup}''.

Some other examples such as ``\texttt{wearing a pink dress/in a pink dress/dressed in pink}'' indicate our SAT-LSTMs model is more robust to syntactic variations, which is more crucial to boost the generalization ability while encoding a sentence or sentence pair.



\subsubsection{Analysis of Learned Sentence Representations}

We explore the sentence representations learned by the three different models on the SNLI.
Table \ref{tab:nn} illustrates the nearest neighbors of sentence representations learned from NBOW, AT-LSTMs, SAT-LSTMs.

As shown in Table \ref{tab:nn}, NBOW finds a sentence's neighbors with full consideration of lexical paraphrase. While the neighbors returned by SAT-LSTMs are mostly syntactic variations with meaning preserving. For example, for the first sentence
``\texttt{the boy are bare chested}'', NBOW gives the ``\texttt{the men are naked}'' most likely based on
the word pair ``\texttt{bare/naked}'', thereby ignoring the information of ``\texttt{chested}''. However, the sentences given by SAT-LSTMs contain the same meaning with ample ways of expressions, such as ``\texttt{the man is not wearing a shirt}'' and ``\texttt{the man without a shirt}'', which accurately reflect the meaning of ``\texttt{bare chested}''.

Compared with AT-LSTMs, SAT-LSTMs can provide more flexible syntactic expressions. For example,
for the sentence `\texttt{a golden retriever nurses puppies}'', SAT-LSTMs capture this syntactic paraphrase
`\texttt{A nurses B/B is snuggling with A}'', which is difficult for NBOW and AT-LSTMs models.

\section{Related Work}

There has been recent work proposing to incorporate syntax priori into neural network.
\newcite{socher2012semantic} use a recursive neural network model that learns compositional vector representations for phrases and sentences of arbitrary syntactic type and length.
\newcite{tai2015improved} introduce a generalization of the standard LSTM architecture to tree-structured network.
\newcite{bowman2016fast} propose an stack-augmented Parser-Interpreter Neural Network for sentence encoding, which combines parsing and interpretation within a single tree-sequence hybrid model.
These models are designed for representing a sentence in more plausible way, while we want to model the strong interaction of two sentences over tree structure.

More recently, several works have tried to incorporate priori into attention based model.
\newcite{cohn2016incorporating} extend the attentional neural translation model to include structural biases from word based alignment models.
\newcite{gu2016incorporating} incorporate copying mechanism into attention based model to address the OOV problem in a more systemic way for machine translation.
Different with these models, we augment attention model with syntax priori for semantic matching.

Another thread of work is sequential attention models for natural language inference.
\newcite{rocktaschel2015reasoning} propose to use attention model for sentence pair encoding.
\newcite{wang2015learning} extend this model by paying more attention to important word-level matching results.
Compared with these models, we integrate syntax structure into attention matching model, which can match two trees in a plausible way.

\section{Outlook}
Natural language has its underlying syntactic structure, which gives a feasibility to assign attention to tree-structured topologies instead of a flat sequence. Although we just use it in context of natural language inference, the idea of syntax-based attention model can be easily transferred to other tasks for phrase-level alignment, such as neural translation model. When we submit our paper, we find this paper \cite{DBLP:journals}, which proposed tree-to-sequence attention based model for neural machine translation, thereby showing the effectiveness of syntax-based attention mechnism. The major difference is their model is based on word-to-word and word-to-phrase attention (sequence conditioned on tree) whereas our proposed model focus on phrase-to-phrase attention (tree over tree).

\section{Conclusion}
In this paper, we integrate syntax structure into attention model. Compared with sequence-based attention model, our model can easily capture phrase-level alignment.
Experiments on Stanford Natural Language Inference Corpus demonstrate the efficacy of our proposed model and its superiority to competitor models.
Furthermore, we have made an elaborate experiment design and case analysis to evaluate the effectiveness of our syntax-base matching model and explain why attention over trees is a good idea.

In future, we wish to use our SAT-LSTMs matching model to learn the representation of phrasal\cite{wieting2015towards} or syntactic paraphrases from massive paraphrase dataset, such as PPDB \cite{ganitkevitch2013ppdb}. We expect that the learned representation of subtree with rich prior knowledge should be useful for downstream tasks in a pre-trained manner.

\bibliographystyle{emnlp2016}
\bibliography{../nlp,../ours,nlp}

\end{document}